\documentclass{article}
\usepackage{ijcai19}

\usepackage{times}
\usepackage{soul}
\usepackage{url}
\usepackage[hidelinks]{hyperref}
\usepackage[utf8]{inputenc}
\usepackage[small]{caption}
\usepackage{graphicx}
\usepackage{amsmath}
\usepackage{booktabs}
\usepackage{enumitem}
\usepackage{amsfonts}
\usepackage{amsmath}
\usepackage{amssymb}
\usepackage{mathtools}
\usepackage{amsthm} 
\usepackage{bm}
\usepackage{graphicx}
\usepackage{subfigure}
\usepackage{multirow}
\usepackage[table,xcdraw]{xcolor}
\usepackage{makecell}
\usepackage[figuresright]{rotating}
\usepackage{balance} 

\usepackage{color}
\definecolor{brown}{RGB}{139,64,0}
\definecolor{pink}{RGB}{255,192,203}

\title{Generative  Diffusion Models on Graphs: Methods and Applications}

\author{
Chengyi Liu$^1$\and
Wenqi Fan$^1$\thanks{Wenqi Fan is corresponding author.}\and
Yunqing Liu$^1$\and
Jiatong Li$^1$\and \\
Hang Li$^2$\and
Hui Liu$^2$\and
Jiliang Tang$^2$\And
Qing Li$^1$\\
\affiliations
$^1$The Hong Kong Polytechnic University, $^2$Michigan State University\\
\emails
wenqifan03@gmail.com,
\{chengyi.liu, yunqing617.liu, jiatong.li\}@connect.polyu.hk,
\\
\{lihang4, liuhui7, tangjili\}@msu.edu,
csqli@comp.polyu.edu.hk
}

\begin{document}

\maketitle

\begin{abstract}

Diffusion models, as a novel generative paradigm, have achieved remarkable success in various image generation tasks such as image inpainting, image-to-text translation, and video generation. Graph generation is a crucial computational task on graphs with numerous real-world applications. It aims to learn the distribution of given graphs and then generate new graphs. Given the great success of diffusion models in image generation, increasing efforts have been made to leverage these techniques to advance graph generation in recent years.  In this paper, we first provide a comprehensive overview of generative diffusion models on graphs, In particular, we review representative algorithms for three variants of graph diffusion models, i.e., Score Matching with Langevin Dynamics (SMLD), Denoising Diffusion Probabilistic Model (DDPM), and Score-based Generative Model (SGM). Then, we summarize the major applications of generative diffusion models on graphs with a specific focus on molecule and protein modeling. Finally, we discuss promising directions in generative diffusion models on graph-structured data. For this survey, we also created a GitHub project
website by collecting the supporting resources for generative diffusion models on graphs, at the link: \href{https://github.com/ChengyiLIU-cs/Generative-Diffusion-Models-on-Graphs}{https://github.com/ChengyiLIU-cs/Generative-Diffusion-Models-on-Graphs} 

\end{abstract}

\section{Introduction}
\label{Introduction}

Graphs can represent the rich variety of relationships (\textit{i.e.,} edges) between real-world entities (\textit{i.e.,} nodes). They have been widely used in a diversity of domains~\cite{ma2021deep,xia2021graph,kinderkhedia2019learning}, such as social networks~\cite{fan2019deep_dscf,derr2020epidemic,fan2019deep}, molecular graph structure~\cite{wu2022diffusionbased}, and recommender systems ~\cite{fan2022graph,fan2020graph}, aiming to model association information and structural patterns among various real-world objects~\cite{barabasi2013network,zhu2022survey}. 
Due to the prevalence of graphs,  graph generative models, with the goal of learning the given graph distributions and generating novel graphs, have attracted significant attention in various applications~\cite{zhang2020deep,faez2021deep}, such as drug discovery and semantic parsing in NLP. 
Typically, there are two graph generation patterns for most existing methods, autoregressive generation and one-shot generation~\cite{jo2022score,zhu2022survey}. 
Particularly, autoregressive generation methods are designed to generate desired graphs in a sequential process, while one-shot generation methods generate the entire graph with topology structure and node/edge feature in one single step. 
In general, graph generation faces three fundamental challenges - (1) \emph{Discreteness}: The graph structure is naturally discrete, resulting in calculation difficulties of models' gradients~\cite{guo2020systematic,zhang2020deep}.
To this end, the most widely used optimization algorithms cannot be straightforwardly introduced to the back-propagation training for graph generation in an end-to-end manner.
(2) \emph{Complex Intrinsic Dependencies}: Unlike image data, nodes are not independent and identically distributed (or i.i.d.). 
In other words, these complex graph structural dependencies are inherent relationships among instances (e.g., nodes and edges)~\cite{niu2020permutation,guo2020systematic}. 
Such complexity of graph structure introduces tremendous challenges in generating desired graphs.  
(3) \emph{Permutation Invariant}:  
Since nodes are naturally unordered in most graphs, there are at most \(N!\) different equivalent adjacency matrices representing the same graph with $N$ nodes~\cite{niu2020permutation}.

Traditional graph generation methods rely on leveraging hand-crafted graph statistics (e.g., degrees and clustering coefficients properties), and learning kernel functions or engineered features to model the structural information  ~\cite{hamilton2017representation}. 
Driven by recent advances in Deep Neural Networks (DNNs) techniques, deep generative models, such as variational autoencoder (VAE)~\cite{simonovsky2018graphvae}, Generative Adversarial Networks (GAN)~\cite{de2018molgan,liu2019learning}, and normalizing flows~\cite{pmlr-v119-kohler20a,luo2021graphdf}, have largely improved the generation performance for graph-structured data. 
For example, GraphVAE estimates the graph distribution by constructing two graph neural networks (GNNs) as an encoder and a decoder~\cite{simonovsky2018graphvae}, and MolGAN introduces a GAN-based framework for molecular generation~\cite{de2018molgan}.   
Although these deep generative methods have achieved promising performance, most of them still have several limitations. 
For example, VAE approaches struggle with the estimation of posterior to generate realistic large-scale graphs and require expensive computation to achieve permutation invariance because of the likelihood-based method~\cite{bond2021deep}. 
Most GAN-based methods are more prone to mode collapse with graph-structured data and require additional computation to train a discriminator~\cite{de2018molgan,wang2018graphgan}. 
The flow-based generative models are hard to fully learn the structural information of graphs because of the constraints on the specialized architectures~\cite{cornish2020relaxing}.
Thus, it is desirable to have a novel generative paradigm for deep generation techniques on graphs.

In recent years, denoising diffusion models have become an emerging generative paradigm to enhance generative capabilities in the image domain~\cite{cao2022survey,yang2022diffusion}.
More specifically, inspired by the theory of non-equilibrium thermodynamics, the diffusion generative paradigm can be modelled as  Markov chains trained with variational inference~\cite{yang2022diffusion},  consisting of two main stages, namely, a forward diffusion and a reverse diffusion. 
The main idea is that they first develop a noise model to perturb the original input data by adding noise (i.e., generally Gaussian noise) and then train a learnable reverse process to recover the original input data from the noise. 
Enhanced by the solid theoretical foundation, the probabilistic parameters of the diffusion models are easy-to-tractable, making tremendous success in a broad range of tasks~\cite{cao2022survey,yang2022diffusion} such as image generation, text-to-image translation, molecular graph modeling.

Recent surveys on deep diffusion models have focused on the image domain~\cite{cao2022survey,yang2022diffusion,croitoru2022diffusion}. Therefore, in this survey, we provide a comprehensive overview of the advanced techniques of deep graph diffusion models. More specifically, we first briefly introduce the basic ideas of the deep generative models on graphs along with three main paradigms in diffusion models.
Then we summarize the representative methods for adapting generative diffusion methods on graphs. 
After that, we systematically present two key applications of diffusion models, i.e., molecule generation and protein modeling. 
At last, we discuss the future research directions for diffusion models on graphs. 
To the best of our knowledge, this survey is the very first to summarize the literature in this novel and fast-developing research area.

\section{Preliminaries}
\label{Preliminaries}

\begin{figure}[t]
\vskip -0.15in
\centering
{\subfigure[Variational AutoEncoders]
{\includegraphics[width=0.68\linewidth]{{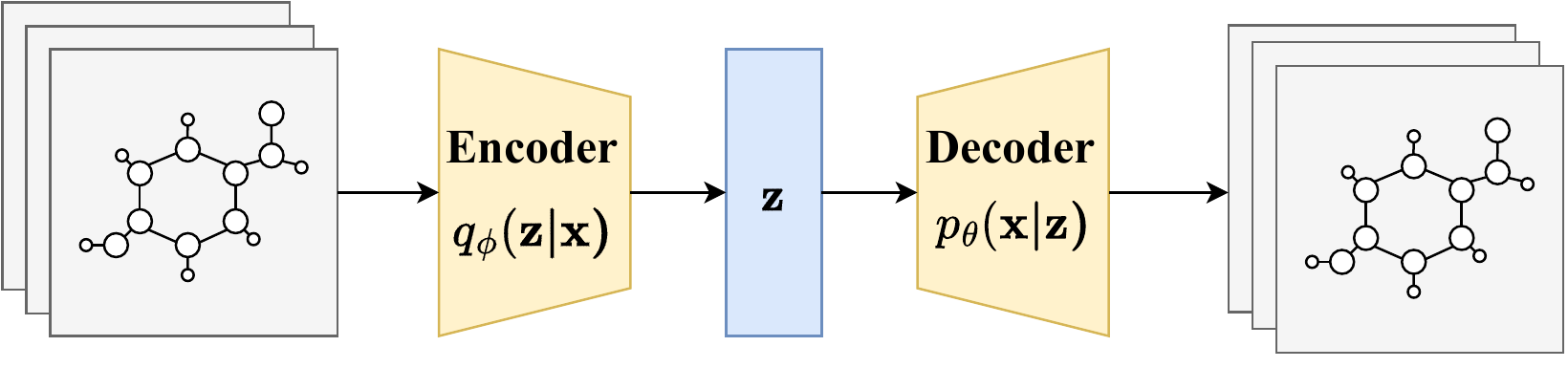}}\label{fig:ciao_att_mae}}}

\vskip -0.051in
{\subfigure[Generative Adversarial Networks (GAN)]
{\includegraphics[width=0.68\linewidth]{{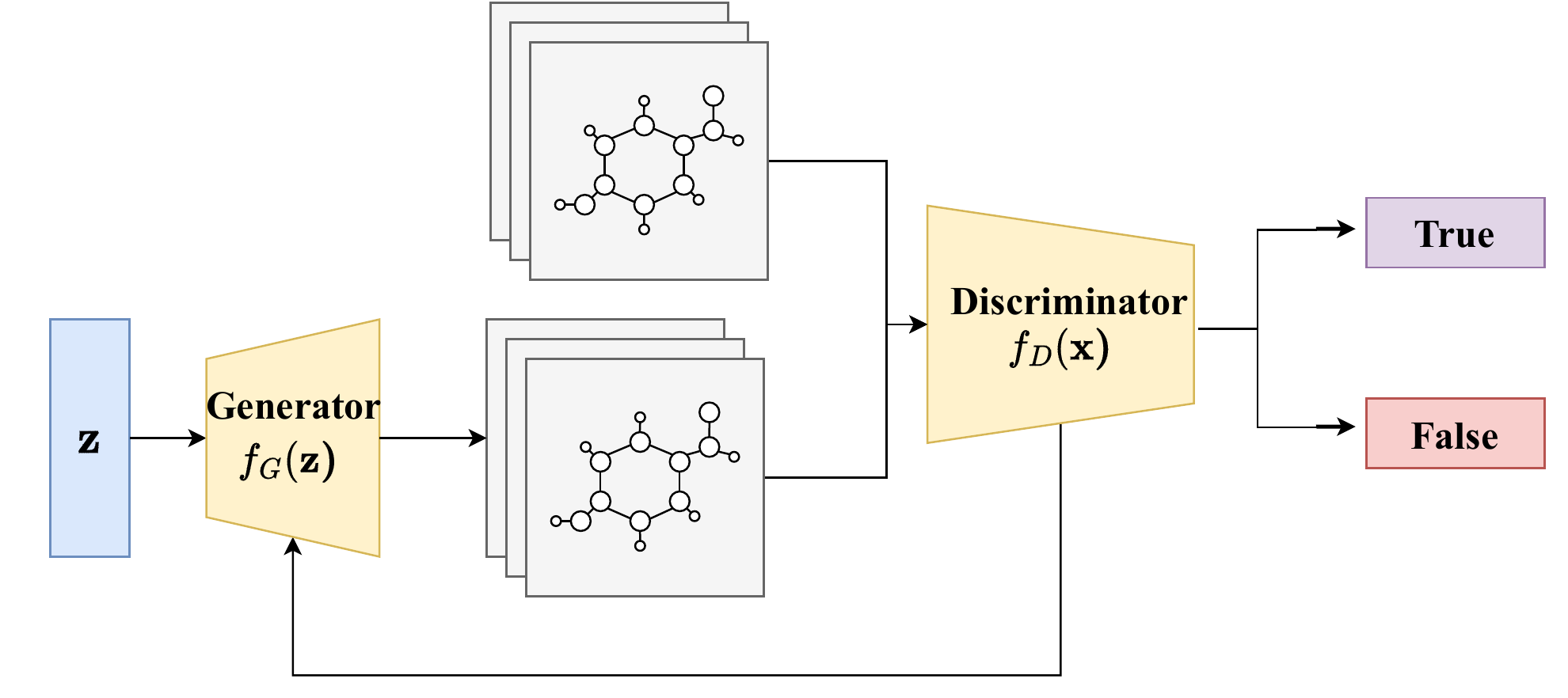}}\label{fig:epinions_att_mae}}}

\vskip -0.051in
{\subfigure[Normalizing Flows]
{\includegraphics[width=0.68\linewidth]{{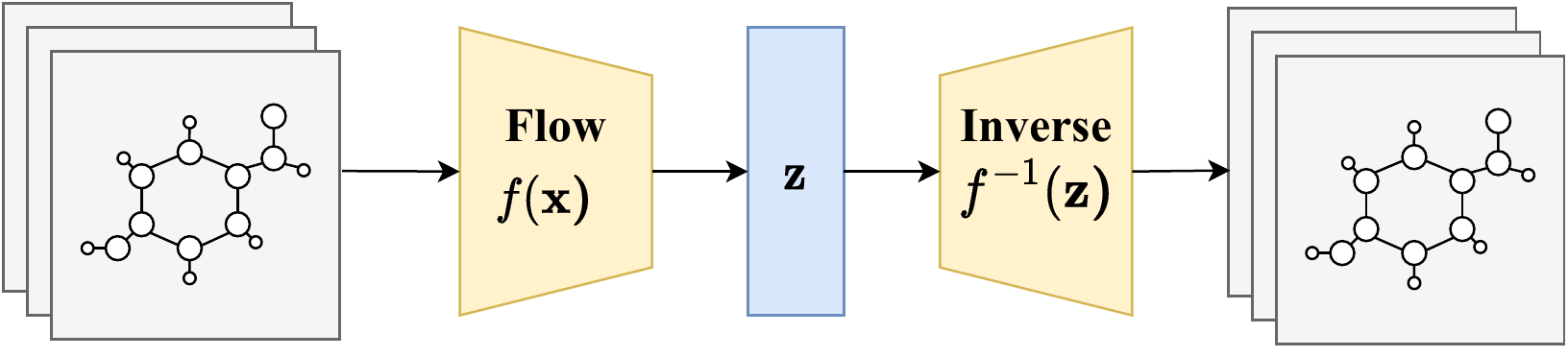}}\label{fig:ciao_att_mae}}}

\vskip -0.051in
{\subfigure[Diffusion models]
{\includegraphics[width=0.82\linewidth]{{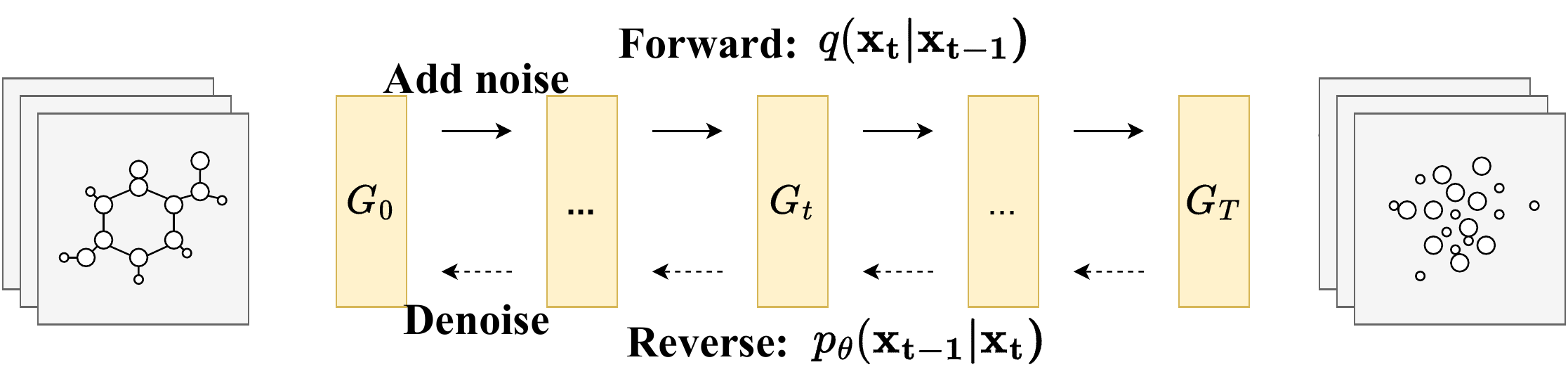}}\label{fig:epinions_att_mae}}}
\vskip -0.15in
\caption{Deep Generative Models on Graphs.
\label{fig:deepGGM}}
\vskip -0.18in
\end{figure}

In this section, we briefly introduce some related work about deep generative models on graphs and detail three representative diffusion frameworks (i.e., SMLD, DDPM, and SGM). The general architecture of these deep generative models on graphs is illustrated in Figure~\ref{fig:deepGGM}.  
Next, we first introduce some key notations.

\noindent \textbf{Notations}.
In general, a graph is represented as $\mathbf{G} \!=\!(\mathbf{X},\!\mathbf{A})$, consisting of  $N$ nodes.
  $\mathbf{A} \!\in\! \mathbb{R}^{\!N\!\times\! N}$ is the adjacency matrix, 
  where $\mathbf{A}_{ij} = 1$  when node $v_i$ and node $v_j$ are connected,  and 0 otherwise.
 $\mathbf{X} \!\in\! \mathbb{R}^{\!N\!\times\! d}$ denotes the node feature with dimension $d$. 
Under diffusion context, $\mathbf{G}_0$ refers to the original input graph, while $\mathbf{G}_t$ refers to the noise graph at the $t$ time step.

\subsection{Deep Generative Models on Graphs}

\noindent \textbf{Variational Autoencoders (VAEs)}.
As the very first deep generative model, variational autoencoders have been successfully applied to graphs, where VAE aims to train a probabilistic  graph encoder $q_{\phi}(\mathbf{z}|\mathbf{G})$ to map the graph space to a low-dimensional continuous embedding $\mathbf{z}$, and a  graph decoder $p_{\theta}(\mathbf{G}|\mathbf{z})$ to reconstruct new data given the sampling from $\mathbf{z}$ ~\cite{kipf2016variational,simonovsky2018graphvae}.

\noindent \textbf{Generative Adversarial Networks (GAN)}.
GAN is to implicitly learn the graph data distribution with the min-max game theory ~\cite{maziarka2020mol,wang2018graphgan,fan2020deep,wang2020global} with two deep neural networks: generator $f_{G}$ and discriminator $f_{D}$.
Specifically, the generator attempts to learn the graph distribution and generate new graphs, while the discriminator tries to distinguish the real graph from the generated graph. 
Due to the discrete nature of graphs, most GAN-based methods are optimized by reinforcement learning techniques.

\noindent \textbf{Normalizing Flows}. 
The normalizing flow leverages a sequence of invertible functions $f(\mathbf{x})$ to map the graph samples (i.e., adjacency matrices and/or edge features) to latent variables $\mathbf{z}$ and learns the graph distribution by tracking the change of density with Jacobian matrix ~\cite{liu2019graph}. The inverse function $f^{-1}(\mathbf{z})$ yields new samples from latent variables by reversing mapping $f(\mathbf{x})$. 
The function $f$ specifies an expressive bijective map which supports a tractable computation of the Jacobian determinant ~\cite{kobyzev2020normalizing}. 
Generally, the training process would estimate the log-likelihoods of each graph sample and update the parameter of $f^{-1}(\mathbf{z})$ by maximizing log-likelihoods with gradient descent.

\noindent \textbf{Limitations}. 
Despite the great success,  most existing  deep generative models are still facing challenges in graph generation.
For instance,  VAE models generate graphs based on likelihood, which requires a massive graph-matching process or an explicit estimation of the likelihood of each possible node alignment when achieving permutation invariant ~\cite{simonovsky2018graphvae}.
In practice, GAN-based  generative models on graphs easily fall into mode collapse, which can limit both the scale and novelty of generated graphs~\cite{de2018molgan}. 
As for normalizing flow, the bijective model structure limits its ability to capture large-scale node-edge dependencies~\cite{cornish2020relaxing}.

\subsection{Diffusion Models}

In general, there are three paradigms of diffusion models: \emph{Score Matching with Langevin Dynamics} (\textbf{SMLD}), \emph{Denoising Diffusion Probabilistic Model} (\textbf{DDPM}),  and \emph{Score-based Generative Model} (\textbf{SGM}). SMLD and DDPM leverage the score matching idea and nonequilibrium thermodynamics respectively to learn different reverse functions of the diffusion process. SGM generalizes the discrete diffusion steps into the continuous scenario and further models the diffusion process with the Stochastic
Differential Equations (SDE). In the following, we will introduce each paradigm in details.

\subsubsection{Score Matching with Langevin Dynamics (SMLD)}

As the first representative version of the diffusion model, SMLD ~\cite{song2019generative} proposes a novel generative model mechanism
that first progressively adds random noise to the data distribution to a predefined prior (usually Gaussian noise), and then reverses the diffusion process by learning the gradient of the data distribution $\nabla_{\mathbf{x}}\log p(\mathbf{x})$. The SMLD perturbs the original distribution with a sequence of random Gaussian noises of incremental scales that can be modelled as $q_{\sigma}(\tilde{\mathbf{x}}|\mathbf{x}):= \mathcal{N}\left(\tilde{\mathbf{x}}|{\mathbf{x}}, \sigma^{2} I\right)$. 
This noise scheme facilitates an accurate score matching by preventing the noised distribution from a low dimensional manifold and providing sufficient training data in low data density regions with large-scale noise. 
The SMLD proposes a Noise Conditional Score Network (NCSN) $s_{\theta}(\mathbf{x}_{t}, \sigma)$ to jointly approximate the score. 
With the annealed Langevin dynamics, the NCSN is able to yield new samples by a gradual denoising process from the Gaussian distribution.

\subsubsection{Denoising Diffusion Probabilistic Model (DDPM)} 

Enhanced by the variational inference, the denoising diffusion probabilistic model (DDPM) ~\cite{ho2020denoising} constructs two parameterized Markov chains to diffuse the data with predefined noise and reconstruct the desired samples from the noise. In the forward chain, the DDPM gradually perturbs the raw data distribution $\left. \mathbf{x}_{0} \right.\sim q\left( \mathbf{x}_{0} \right)$ to converge to the standard Gaussian distribution $\mathbf{z}_t$ under a pre-designed mechanism. Meanwhile, the reverse chain seeks to train a parameterized Gaussian transition kernel to recover the unperturbed data distribution. 
Mathematically, the definition of the forward process $q$ is as follows:
\vskip -0.10in
\begin{equation}
\hspace*{-3.2mm}
\fontsize{8pt}{8pt}\selectfont
\begin{aligned}
q\left( {\mathbf{x}_{t}| \mathbf{x}_{t - 1}} \right) &= \mathcal{N}\left( \mathbf{x}_{t};\sqrt{1 - \beta_{t}}\mathbf{x}_{t - 1},\beta_{t}I \right),\\
q(\mathbf{x}_{1:T}|\mathbf{x}_0) &= \prod_{t=1}^{T} q(\mathbf{x}_t| \mathbf{x}_{t-1})\label{DDPM:forward},
\hspace*{-5mm} 
\end{aligned}
\fontsize{10pt}{10pt}\selectfont
\end{equation}
where \(\beta_{t}\in(0,1)\) represents the variance of the Gaussian noise added at time step \textit{t}. 
With $\alpha_{t} = 1 - \beta_{t},~{\overset{-}{\alpha}}_{t} = ~{\prod_{i = 1}^{t}\alpha_{i}}$,  the marginal can be written as:
\begin{equation}
\hspace*{-3.2mm}
\fontsize{8pt}{8pt}\selectfont
\begin{aligned}
q(\mathbf{x}_t | \mathbf{x}_0) &= \mathcal{N}(\mathbf{x}_t; \sqrt{\bar{\alpha}_t} \mathbf{x}_0, (1-\bar{\alpha}_t) I), \\
\mathbf{x}_t &=\sqrt{\bar{\alpha}_t} \mathbf{x}_0 + \sqrt{1-\bar{\alpha}_t} \epsilon, 
\hspace*{-5mm} 
\end{aligned}
\fontsize{10pt}{10pt}\selectfont
\end{equation}
where $\epsilon$ denotes the Gaussian noise. These equations enable the DDPM to sample the noised latent $\mathbf{x}_t$ at an arbitrary step conditioned on $\mathbf{x}_0$ \cite{vignac2022digress}. Meanwhile, the reverse Gaussian transitions $p_{\theta}$ parameterized by $\theta$ can be defined as: 
\vskip -0.250in
\begin{equation}
\hspace*{-3.2mm}
\fontsize{8pt}{8pt}\selectfont
\begin{aligned}
p_\theta(\mathbf{x}_{0:T})&=
p( \mathbf{x}_{T}\prod_{t=1}^{T}p_\theta( \mathbf{x}_{t-1}|  \mathbf{x}_t),\\
p_\theta( \mathbf{x}_{t-1} | \mathbf{x}_t) &= \mathcal{N}( \mathbf{x}_{t-1}; \mu_{\theta}( \mathbf{x}_t, t), \Sigma_{\theta}( \mathbf{x}_t, t)).
\hspace*{-5mm} 
\end{aligned}
\fontsize{10pt}{10pt}\selectfont
\end{equation}
The neural network will be trained to optimize the variational upper bound on negative log-likelihood, which can be estimated via the Monte-Carol algorithm \cite{vignac2022digress}. As a result, the DDPM would sample from the limit distribution, and then recursively generate samples $\mathbf{x}_t$ using the learned reverse chain.

\subsubsection{Score-based Generative Model (SGM)}

The score SDE formula describes the diffusion process in continuous time steps with a standard Wiener process. The forward diffusion process in infinitesimal time can be formally represented as ~\cite{song2020score}:
\vskip -0.08in
\begin{equation}
    \mathrm{d} \mathbf{x} = {f}(\mathbf{x}, t) \mathrm{d} t + {g}(t) \mathrm{d} \mathbf{w}, 
    \label{SDE:forward}
\end{equation}
\vskip -0.08in
\noindent
where $\mathbf{w}$ denotes a standard Wiener process (a.k.a., Brownian motion), and ${g}(\cdot)$ denotes the diffusion coefficient, which is supposed to be a scalar independent of $\mathbf{x}$. The reserve-time SDE describes the diffusion process running backwards in time to generate new samples from the known prior $\mathbf{x}_T$, which is shown as follows: 
\vskip -0.20in
\begin{equation}
    \mathrm{d}\mathbf{x} = [f(\mathbf{x}, t) - g(t)^2  \nabla_{\mathbf{x}}  \log p_t(\mathbf{x})] \mathrm{d} t + g(t) \mathrm{d} \bar{\mathbf{w}}.
    \label{SDE:reverse}
\end{equation}
\vskip -0.08in
The only unknown information in reserve-time SDE is the score function $\nabla_{\mathbf{x}} \log p(\mathbf{x})$, which can be approximated by a time-dependent score-based model ${s}_{\theta}(\mathbf{x}, t)$ by optimizing denoising score-matching objective:
\vskip -0.15in
\begin{equation}
\mathbb{E}_{t,\mathbf{x}_{0},\mathbf{x}_{t}}\left\lbrack \lambda(t)\left| \left| {s_{\theta}\left( \mathbf{x}_{t},t \right) - {\nabla\ }_{\mathbf{x}_{t}}{\log{{p}_{0t}\left( \mathbf{x}_{t} \middle| \mathbf{x}_{0} \right)}}} \right| \right|^{2} \right\rbrack,
\end{equation}
\vskip -0.10in
\noindent
where ${p}_{0t}(\mathbf{x}_{t} | \mathbf{x}_{0})$ represents the probability distribution of $\mathbf{x}_{t}$ conditioned on $\mathbf{x}_{0}$, $t$ is uniformly sampled over $[0, T]$, $\mathbf{x}_{0} \sim p_0(\mathbf{x})$ and $\mathbf{x}_t \sim p_{0t}(\mathbf{x}_t | \mathbf{x}_0)$. The score-based diffusion model through SDE unifies the SMLD and DDPM into a continuous version. 

Furthermore, the probability flow ODE ~\cite{song2020score} designs a deterministic reverse-time diffusion process, whose marginal probability density is identical to the SDE formula. This method largely accelerates the sampling process considering that it allows to perform Gaussian sampling at adaptive time intervals with discretization strategy rather than at successive time steps, and thus the number of estimations of score function is reduced. The reverse-time ODE is defined as below: 

\centerline{$ 
\mathrm{d} \mathbf{x} = [{f}(\mathbf{x}, t) - \frac{1}{2} g(t)^2\nabla_{\mathbf{x}}\log p_t(\mathbf{x})] \mathrm{d} t. 
$}

\newcommand{\pscore}[2]{\nabla_{\!\!#1}\!\log p_t(#2)}
\newcommand{\overbar}[1]{\mkern 1.5mu\overline{\mkern-1.5mu#1\mkern-1.5mu}\mkern 1.5mu}
\newcommand{\dscore}[2]{\nabla_{\!\!\mathbf{#1}_t}\!\log p_{0t}(\mathbf{#2}_t|\mathbf{#2}_0)}

\section{Generative Diffusion Models on Graphs}
\label{Graph Generative Diffusion Models}

In this section, we will categorize existing diffusion techniques on graph-structured data into the three paradigms we introduced above.

\subsection{SMLD on Graphs}

EDP-GNN~\cite{niu2020permutation} is the very first score matching based diffusion method for undirected graph generation. 
Through modeling the symmetric adjacency matrices regarding the different scales of Gaussian noise added to the upper triangular segment with neural network, EDP-GNN learns the score of the graph distribution.
By using a similar annealed Langevin dynamics implementation as SMLD~\cite{song2019generative}, the adjacency matrices are generated from the sampled Gaussian noise. 
Inspired by the GIN method~\cite{xu2018how}, EDP-GNN also introduces a multi-channel GNN layer to obtain node features with the message-passing mechanism and a MLP output layer including a noise-conditioned term to prevent separately training the score network at each noise scale. ConfGF~\cite{molecular_conformation_generation_shi_21} is the first work adapting the SMLD-based diffusion work to the molecular confirmation generation problem. Unlike EDP-GNN, whose target is to generate adjacency matrices, ConfGF focuses on generating atomic coordinates (node feature) $\mathbf{R}$ given the molecular graph \(\mathbf{G}\). Due to the roto-translation equivalent property, ConfGF maps a set of atomic coordinates to a set of interatomic distances $l$. 
By injecting the Gaussian noise over $l$, ConfGF learns the score function of interatomic distance distributions. Similar to EDP-GNN, the score function is later combined with the annealed Langevin dynamics to generate new atomic coordinate samples.

\subsection{DDPM on Graphs}
The adaption of denoising diffusion probabilistic models on graphs is mainly focusing on designing the appropriate transition kernel of the Markov chain. The previous diffusion models usually embed the graphs in continuous space, which might lead to structural information loss. ~\citeauthor{haefeli2022diffusion} propose a denoising diffusion kernel to discretely perturb the data distribution. 
At each diffusion step, each row of the graphs' adjacency matrices is encoded in a one-hot manner and multiplied with a double stochastic matrix $\mathbf{Q}_t$. In the reverse process, the model includes a re-weighted ELBO as the loss function to obtain stable training.
With discrete noise, the sampling process is largely accelerated.  
Furthermore, DiGress~\cite{vignac2022digress} extends the DDPM algorithm to generate graphs with categorical node and edge attributes. 
The conditional probabilities for the noisy graphs can be defined as follows:
\vskip -0.1in
\begin{equation}
\hspace*{-3.2mm}
\fontsize{8pt}{8pt}\selectfont
\begin{aligned}
     q(\mathbf{G}_t | \mathbf{G}_{t-1}) &= ({\mathbf{X}}_{t-1} {\mathbf{Q}}^\mathbf{X}_t, {\mathbf{E}}_{t-1} {\mathbf{Q}}^\mathbf{E}_t), \\
     q(\mathbf{G}_t | \mathbf{G}) &= ({\mathbf{X}} \bar{\mathbf{Q}}^\mathbf{X}_t, \mathbf{E} \bar {\mathbf{Q}}^\mathbf{E}_t),  
\label{discrete diffuse}
\hspace*{-5mm} 
\end{aligned}
\fontsize{10pt}{10pt}\selectfont
\end{equation}
where $\mathbf{G}_t = (\mathbf{X}_t, \mathbf{E}_t)$ refers to the noisy graph composed of the node feature matrix $\mathbf{X}_t$ and the edge attribute tensor $\mathbf{E}_t$ at step $t$. $\mathbf{Q}^{\mathbf{X}}_t$ and $\mathbf{Q}^{\mathbf{E}}_t$ refer to the noise added to the node and edge, respectively. This Markov formulation allows sampling directly at an arbitrary time step without computing the previous steps. 
In the denoising process, DiGress incorporates the cross-entropy to evaluate the distance between the predicted distribution and the input graph distribution with respect to node and edge, so as to train the parameterized graph transformer network $\phi_{\theta}$. 
Thus, the modeling of graph distribution is simplified to a sequence of classification.
In addition, operating on discrete steps allows DiGress to leverage various graph descriptors, such as spectral features, to guide the diffusion process. Overall, DiGress is capable of yielding realistic large-scale graphs depending on the overall or partial graph. Moreover, the E(3) Equivariant Diffusion Model (EDMs) is able to operate on both continuous and categorical features of the graph by training an equivariant network ~\cite{EDM}. The EDMs jointly inject the Gaussian noise to the latent variables $\mathbf{z}_t = [\mathbf{z}^\mathbf{x}_t, \mathbf{z}^\mathbf{h}_t]$ of nodes coordinates (continuous) $\mathbf{x}_i$ and the other features $\mathbf{h}$ (categorical). As an extension of EDMs, Equivariant Energy-Guided SDE (EEGSDE) \cite{EEGSDE} introduces a novel property prediction network that serves as a guiding mechanism in the generative process. This network is concurrently trained alongside the reverse diffusion process, with its gradient serving as an additional force to enhance the overall performance.
Current generative models struggle to effectively capture the complexities of interatomic forces and the presence of numerous local constraints. To address this issue, the proposed approach in MDM \cite{MDM} utilizes augmented potential interatomic forces and incorporates dual equivariant encoders to effectively encode the varying strengths of interatomic forces. 
Additionally, a distributional controlling variable is introduced  to ensure thorough exploration and enhance generation diversity during each diffusion/reverse step.
Although the diffusion method is initially designed for a one-shot generative manner, the GRAPHARM model proposes an autoregressive diffusion model to generate graphs by sequentially predicting each row in the adjacency matrix ~\cite{anonymous2023autoregressive}. The GRAPHARM masks nodes and corresponding edges in forward diffusion whose order is determined by a diffusion ordering network $q_{\phi}({\sigma}|\mathbf{G}_0)$. In the reverse process, the GRAPHARM denoises only one node at each step (i.e., sequentially generates one row in the adjacency matrix) with the help of  graph attention networks ~\cite{liao2019efficient}.

\subsection{SGM on Graphs}
Although EDP-GNN develops a score-based generative model to derive the adjacency matrix of the graph, the estimation for the score function depends on the noise scales at the discrete steps, which restricts its capacity to produce large-scale graphs. 
GraphGDP ~\cite{huang2022graphgdp} leverages the variance-preserving SDE to disturb the adjacency matrix to random graphs. In the reverse process, the Position-enhanced Graph Score Network (PGSN) incorporates the features of both nodes and edges and graph position information for permutation equivariant score estimation. Notably, GraphGDP defines a transformation in the forward process to associate continuous distribution with discrete graphs, which allows the model to learn additional graph information of intermediate diffusion steps. Moreover, 
GDSS ~\cite{jo2022score} proposes a continuous-time SDE system to model the diffusion process over nodes and edges simultaneously,  where Gaussian noise is directly added to the adjacency matrix and node features.
Formally, the forward diffusion process on the weighted graph $\mathbf{G}$ at each infinitesimal time step can be modelled as follows: 
\begin{equation}
\fontsize{9.5pt}{9.5pt}\selectfont
\begin{aligned}
    \mathrm{d}\mathbf{G}_t = f_t(\mathbf{G}_t)\mathrm{d}t + g_t(\mathbf{G}_t)\mathrm{d}\mathbf{w} , \;\;\; \mathbf{G}_0\sim p_{data},
    \label{GDSS:forward_diffusion}
\end{aligned}
\fontsize{10pt}{10pt}\selectfont
\end{equation}
where $f_t$ represents the linear drift coefficient.
To reduce the computation in the reverse diffusion process, GDSS introduces a reverse-time SDE system with respect to nodes and edges as follows:
\begin{align}
\fontsize{9pt}{9pt}\selectfont
\hspace*{-3mm}
\begin{cases}
    \! \mathrm{d}\mathbf{X}_t \! = \! \left[{f}_{1,t}(\mathbf{X}_t) - g_{1,t}^2\! \pscore{\mathbf{X}_t}{\mathbf{X}_t,\! \mathbf{A}_t} \right]\! \mathrm{d}\overbar{t} + g_{1,t}\mathrm{d}\bar{\mathbf{w}}_{1}, \\[5pt]
    \! \mathrm{d}\mathbf{A}_t = \! \left[{f}_{2,t}(\mathbf{A}_t) - g_{2,t}^2\! \pscore{\mathbf{A}_t}{\mathbf{X}_t,\! \mathbf{A}_t} \right]\! \mathrm{d}\overbar{t} + g_{2,t}\mathrm{d}\bar{\mathbf{w}}_{2},
\end{cases}\label{GDSS:reverse_diffusion}
\fontsize{9pt}{9pt}\selectfont
\hspace{-5mm}
\end{align}
where ${\nabla_{\mathbf{X}_t}\!\log p_t(\mathbf{X}_t,\! \mathbf{A}_t)}$ and ${\nabla_{\mathbf{A}_t}\!\log p_t(\mathbf{X}_t,\! \mathbf{A}_t)}$ are the partial score functions, which refer to the gradients of the joint log-density connecting the adjacency matrix $\mathbf{A}$ and node feature matrix $\mathbf{X}$. 
GDSS also proposes a corresponding object function to jointly estimate the log density of nodes and edges:
\begin{equation}
\hspace*{-3.2mm}
\fontsize{8pt}{8pt}\selectfont
\begin{aligned}
\begin{cases}
    \! \min_{\theta}\mathbb{E}_t\!\! \left\{\! \lambda_1(t) \mathbb{E}_{\mathbf{G}_0}\! \mathbb{E}_{\mathbf{G}_t|\mathbf{G}_0}\!\! \left\| {s}_{\theta,t}(\mathbf{G}_t) - \dscore{X}{X} \right\|^2_2\right\} \\
    \! \min_{\phi}\mathbb{E}_t\!\! \left\{\! \lambda_2(t) \mathbb{E}_{\mathbf{G}_0}\! \mathbb{E}_{\mathbf{G}_t|\mathbf{G}_0}\!\! \left\| {s}_{\phi,t}(\mathbf{G}_t) - \dscore{A}{A} \right\|^2_2\right\}
    \label{GDSS:objective}
\end{cases}
\hspace*{-5mm}
\end{aligned}
\fontsize{10pt}{10pt}\selectfont
\end{equation}
where ${s}_{\theta, t}(\mathbf{G}_t)$ and ${s}_{\phi, t}(\mathbf{G}_t)$ are the MLP advanced by the graph multi-head attention blocks ~\cite{baek2021accurate} to learn the long-term relationship. The GDSS approximates the expectation in Eq.(\ref{GDSS:objective}) with Monte Carlo estimation, which requires fewer computation and sampling steps compared to the Langevin dynamics~\cite{jo2022score}.
To further improve the generation quality, GDSS proposes an integrator to correct the estimated partial score by the score-based MCMC estimation. Note that GDSS is the very first diffusion framework that enables the generation of a whole graph based on node-edge dependency. 
However, the standard diffusion process would eliminate the features of the sparse graphs in a few steps, which may cause the score estimation uninformative in the reverse process. To address such limitation, a Graph Spectral Diffusion Model (GSDM) ~\cite{luo2022fast} is introduced to perform the low-rank Gaussian noise insertion, which can gradually perturb the data distribution with less computation consumption while achieving higher generation quality. 
To be specific, in the diffusion process, GSDM performs spectral decomposition on the adjacency matrix $\mathbf{A}$ to obtain diagonal eigenvalue matrix $\mathbf{\Lambda}$ and eigenvectors $\mathbf{U}$ (i.e., $\mathbf{A} = \mathbf{U} \mathbf{\Lambda} \mathbf{U}^\top$). Meanwhile, since the top-$k$ diagonal entries in $\mathbf{\Lambda}$ can maintain most of the graph information, GSDM conducts the diffusion on the corresponding top-$k$ largest eigenvalues in $\mathbf{A}$ for efficient computation.
In addition, SGGM introduces a latent-based generative framework on the graph~\cite{anonymous2023scorebased}, which first encodes the high-dimensional discrete space to low-dimensional topology-injected latent space via a pre-trained variational graph autoencoder and then adopts score-based generative model for the graph generation.

\section{Applications}
\label{sec:application}

\begin{figure}[t] 
\vskip -0.15in
\centering 
\includegraphics[width=0.99\linewidth]{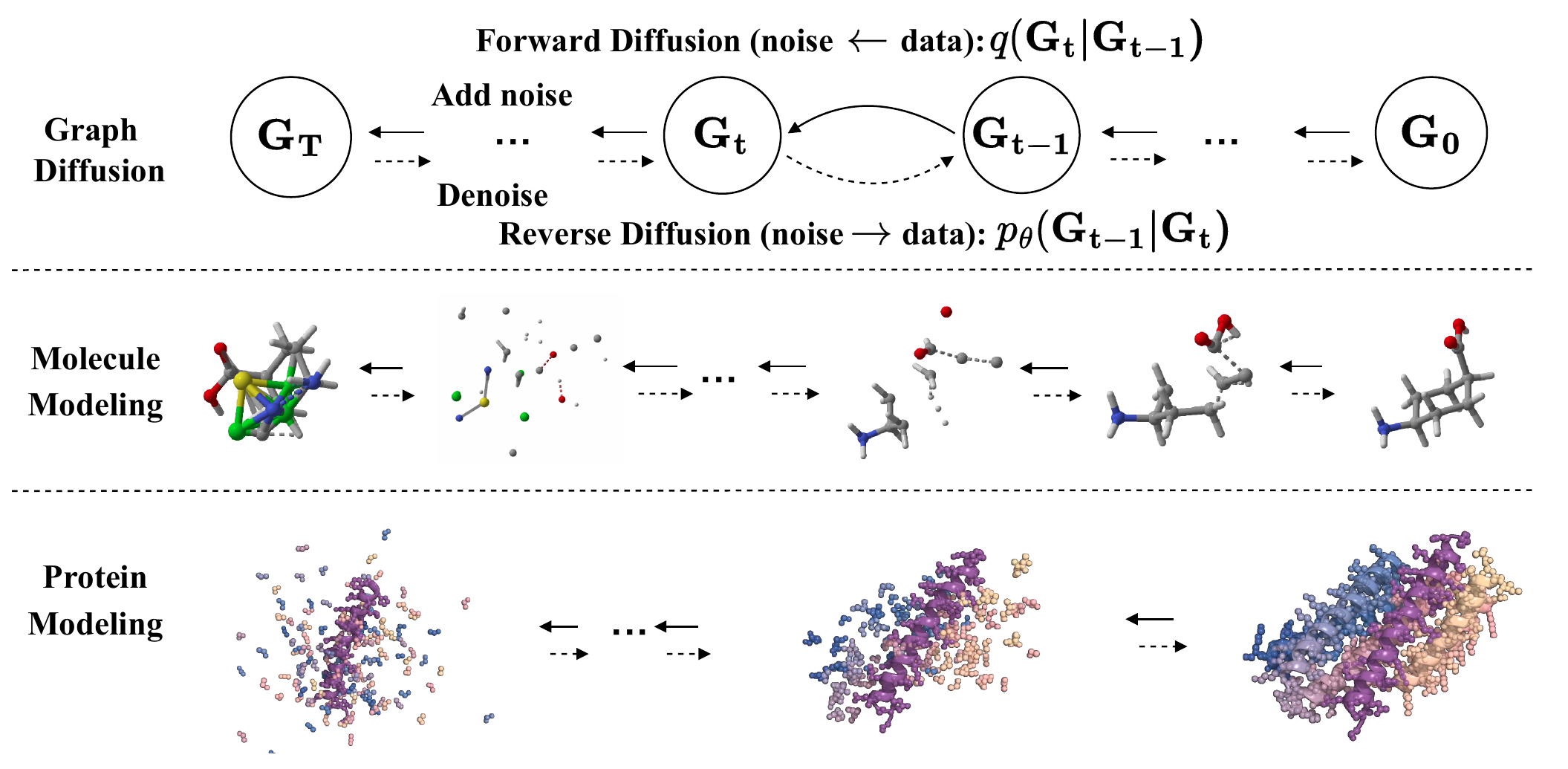}
\vskip -0.12in
\caption{An illustration of diffusion models on molecular and protein generations. 
The forward diffusion process involves the gradual addition of noise from the fixed posterior distribution \(q(\textbf{G}_t|\textbf{G}_{t-1})\) to the input graph \(\textbf{G}_0\) (representing a molecule or protein) over a period of time \(T\) steps, ultimately resulting in the destruction of the molecule or protein structure.
In contrast, the reverse diffusion process samples an initial graph \(\textbf{G}_T\) from a standard Gaussian distribution and gradually refines the graph's structure by using Markov kernels \(p_{\theta}(\textbf{G}_{t-1}|\textbf{G}_T)\).}
\label{Application} 
\vskip -0.15in
\end{figure}

In this section, we aim to review key applications for diffusion models on graphs.
We focus on molecule and protein generations that have been widely used in the chemistry and biology domains. 
In Figure \ref{Application}, we use a generic framework to illustrate diffusion models on molecule and protein generation tasks. 
In Table \ref{tab:application}, we summarize these applications.

\subsection{Molecule Modeling}
The goal of molecule modeling is to employ graph learning techniques for the purpose of representing to better perform downstream tasks, such as molecule generation ~\cite{guo2022graph,sanchez2018inverse}. 
In general, molecules can be naturally represented as graphs (e.g., atom-bond graphs), where atoms and bonds are represented as nodes and edges. 
As such, graph learning techniques can be applied to analyze and manipulate molecular structures for various downstream tasks, such as drug discovery, computational chemistry, materials science,  bioinformatics, etc. 
Furthermore, molecular graph modeling can be used to generate new molecules with desired properties by using advanced graph learning techniques such as VAE~\cite{liu_constrained_2018,GF-VAE,2-stageVAE}, GNNs~\cite{rong_self-supervised_2020,Mercado_2021,han_gnn-retro_2022}, and reinforcement learning (RL)~\cite{zhou2019optimization,olivecrona2017molecular}. 
Particularly, molecule modeling can be further classified into two tasks, namely, molecule conformation generation and molecular docking. 

\subsubsection{Molecule Conformation Generation}

A molecule can be represented by three-dimensional geometry or conformation, in which atoms can be denoted as their Cartesian coordinates. 
The biological and physical characteristics of the molecule are significantly influenced by its three-dimensional structure.
Meanwhile, molecular conformations possess roto-translational invariance. 
As a result, several techniques use intermediary geometric variables that also have roto-translational invariance, such as torsion angles, bond angles and atomic distances, to avoid forthrightly modeling atomic coordinates~\cite{molecular_conformation_generation_shi_21,GeoMol}.
However, as they aim to model the intermediate geometric variables indirectly, they may be subject to limitations in either the training or inference process.
To address this issue, GeoDiff~\cite{xu2022geodiff} treats atoms as thermodynamic system particles and uses nonequilibrium thermodynamics to simulate the diffusion process. 
By learning to reverse the diffusion process, the sampled molecules gradually diffuse backwards into the target conformation. 
Dynamic Graph Score Matching (DGSM)~\cite{luo2021predicting} uses score matching based on a 2D molecular graph to generate the conformation structure of a molecule by modeling  local and long-range interactions of atoms.

Torsional Diffusion~\cite{jing2022torsional} defines the diffusion over a torus to represent torsion angles, providing a more natural parameterization of conformers. 
Besides, Torsional Diffusion leverages the torus as a representation of torsion angles for conformations generation.
By incorporating a probabilistic analysis that enables probability calculations and atom classification, E(3) Equivariant Diffusion Model ({EDMs})~\cite{EDM} enables the model to learn the denoising process in continuous coordinates and improve the generation of molecular conformations.
As an extension of EDMs,  equivariant energy-guided stochastic differential equations ({EEGSDE}) ~\cite{EEGSDE} adds energy functions to the model as a guide, so as to  learn the geometric symmetry of molecules to generate 3D molecular conformations.

In addition to  adding energy guidance and learning the atomic coordinates of molecules and torsion angles, other domain knowledge can be introduced into the model for enhancing  the molecular representation learning.
For example, to model interatomic interactions in molecular representation, MDM~\cite{MDM} considers the role of atomic 
spacing in iteratomic forces for molecular representation. 
Because chemical bonds control interatomic forces when atoms are sufficiently close to one another, MDM treats atomic pairs with atomic spacing below a specific threshold as covalent bonds. 
For atomic pairs with atomic spacing above a certain threshold, the van der Waals force is proposed to govern the interatomic forces.  
Additionally, to enhance the diversity of molecule generation, they introduce latent variables that are interpreted as control representations in each diffusion/reverse step of the diffusion model. 
DiffBridges~\cite{wu2022diffusionbased} designs an energy function with physical information and a statistical prior for molecule generation. It differs from other methods in its incorporation of physical prior into bridges, as opposed to learning diffusions as a combination of forward-time diffusion bridges.

Transformer architectures have achieved notable success in the fields of natural language processing and computer vision, such as ViT~\cite{dosovitskiyimage}, BERT~\cite{kenton2019bert}, GPT~\cite{brown2020language}.
Similarly, graph transformer~\cite{GraphTransformer} is incorporated into discrete diffusion model with discrete diffusion~\cite{DisDiff} for graph generation.
In particular, DiGress uses graph-based architectures and a noise model that maintains the types of nodes and edges' marginal distributions~\cite{vignac2022digress}, rather than using uniform noise as a starting point. 
A denoising network is then augmented with structural features, enabling conditional generation through guidance procedures.

Since most existing molecular generation approaches generate graphs that are likely similar to training samples, Molecular Out-of-distribution Diffusion (MOOD)~\cite{MOOD} includes an out-of-distribution (OOD) control into the generative stochastic differential equation (SDE), to generate a new molecule graph that is distinct from those in the training set. 
GDSS~\cite{jo2022score} learns the underlying distribution of graphs by establishing score matching goals that are specific to the proposed diffusion process, enabling the estimation of the gradient of the joint logdensity w.r.t. each component.

\subsubsection{Molecular Docking}
Molecular docking is a computational task for predicting the preferred orientation of one molecule when bound to a second molecule, usually a protein, to each other.  
It's used in drug discovery to find the best fit of a small molecule into the active site of a target protein.

Autoregressive models are widely adopted to generate 3D molecules for the protein binding pocket ~\cite{shin2021protein}. 
However, autoregressive models might struggle with capturing complex relationships and interactions between residues in the pocket.
To address these challenges, DiffBP~\cite{DiffBP} generates 3D molecular structures in a non-autoregressive manner while satisfying the physical properties of the molecule, based on the protein target as a constraint.
Using diffusion models and SE(3)-equivariant networks, TargetDiff~\cite{Target-aware} learns atomic types and atomic coordinates to generate protein target molecules with satisfying geometric properties.

Fragment-based drug discovery is also a widely adopted paradigm in drug development, which can provide promising solutions for molecular docking by generating 3D molecules fragment-by-fragment and incorporating diffusion models. 
For instance, {FragDiff}~\cite{FragDiff} generates 3D molecules fragment-by-fragment for pockets. 
In each generation step, FragDiff generates a molecular fragment around the pocket. The atom types, atom coordinates and bonds on this fragment are predicted. Then the fragments are gradually joined together to produce the complete molecule.
Given some fragments, {DiffLink}~\cite{DiffLink} generates the rest of a molecule in 3D. 
The generator of DiffLink is an E(3) equivariant denoising diffusion model to generate fragments. It is conditioned on the positions of the fragment atoms, and optionally also on the protein pocket that the molecule should fit into. 
Finally, DiffLink splices these fragments into a complete drug candidate molecule.

Similar to the transformation of sentiment classification tasks into generative tasks in the NLP field ~\cite{liu2023pre}, 
DiffDock~\cite{DiffDock} uses diffusion model to form docking pose prediction problem as a generation problem and executes a reverse diffusion process using separate ligands and proteins as inputs by randomly selecting the initial states and ranking them.

\subsection{Protein Modeling}
Protein modeling is to generate and predict the structure of proteins. This task is instrumental in comprehending the function and interactions of proteins, and is widely used in the fields of drug discovery and the design of novel proteins with specific characteristics. Previously, proteins have been represented as sequences of amino acids, leading to successes in modeling proteins using language models~\cite{ferruz2022protgpt2,coin2003enhanced}. With the advent of diffusion models in image generation~\cite{ramesh2021zero,nichol2021glide}, a growing number of applications using diffusion models in protein modeling have emerged.

\subsubsection{Protein Generation}
The objective of computational protein design is to automate the generation of proteins with specific structural and functional properties. This field has experienced significant advancements in recent decades, including the design of novel 3D folds~\cite{jumper2021highly}, enzymes~\cite{giessel2022therapeutic} and complexes~\cite{reau2023deeprank}.

Pre-training protein representations on enormous unlabeled protein structures have drawn more and more interest from researchers. 
Siamese Diffusion Trajectory Prediction ({SiamDiff})~\cite{SiamDiff} obtains counterparts by adding random structural perturbations to natural proteins, and diffuses them structurally and sequence-wise through the pre-training process. The diffusion process of the original protein and its counterpart is referred to as two related views. In this process, SiamDiff uses the noise of one view to predict the noise of the other view to better learn the mutual information between the two views.
In contrast to pre-training techniques, {ProteinSGM}~\cite{lee_proteinsgm_2022} applies conditional generation to generate proteins by coating plausible backbones and functional sites into structures of predetermined length.  ProSSDG~\cite{ProSeq} combines the protein's structure and sequence to generate proteins with the desired 3D structures and chemical properties. Based on a brief description of the protein's topology, ProSSDG generates a complete protein configuration.

Despite recent developments in protein structure prediction, it is still difficult to directly generate a variety of unique protein structures using DNNs techniques. 
DiffFold~\cite{foldDiff} generates protein backbone structures by imitating natural folding progress.
Particularly, DiffFold develops novel structures from a chaotic unfolded condition to a stable folded shape through denoising. In addition, DiffFold represents the protein backbone structure as a sequential angular sequence representing the relative orientation of the constituent amino acid residues.

A stable protein backbone that has the motif is referred to as a scaffold. It can greatly benefit from building a scaffold that supports a functional motif. 
SMCDiff~\cite{SMCDiff} uses a particle filtering algorithm for conditional sampling of protein backbone structures, where priority is given to backbone structures more consistent with the motif.

Immune system proteins (called antibodies) attach to particular antigens like germs and viruses to defend the host. The complementarity-determining regions (CDRs) of the antibodies play a major role in regulating the interaction between antibodies and antigens. 
DiffAntigen~\cite{luo2022antigenspecific} jointly generates the sequence and structure of the CDRs of an antibody, based on the framework area of antibody and the target antigen.
DiffAntigen is able to regulate the generation at the antigen structure, not just in the framework region. Additionally, it can also predict the side-chain orientation.

RFdiffusion~\cite{RFDiff} combines a diffusion model with a protein prediction model RoseTTAFold~\cite{RF}. 
During the forward diffusion progress, RFdiffusion perturbs the 3D structural coordinates locally to enhance the model's representational capacity.

\subsubsection{Protein-ligand Complex Structure Prediction}
The prevalence of protein-ligand complexes makes predicting their 3D structure valuable for generating new enzymes and drug compounds.
NeuralPLexer~\cite{NeuralPLexer} predicts the structure of protein-ligand complexes by combining multi-scale induced bias in biomolecular complexes with diffusion models. It takes molecular graphs as ligand input and samples 3D structures from a learned statistical distribution. 
To overcome the difficulties of high-dimensional modeling and to extend the range of protein-ligand complexes, DiffEE~\cite{DiffEE} proposes an end-to-end diffusion generative model, which is based on pre-trained protein language model. DiffEE is able to generate a variety of structures for protein-ligand complexes with correct binding pose.

\begin{table}[t]
\vskip -0.12in
\centering
\caption{A summary of representative applications for generative diffusion method on graphs.}
\vskip -0.12in
\label{tab:application}
\renewcommand\arraystretch{1.25}
\scalebox{0.68}
{
\begin{tabular}{cccc}
\toprule

\multicolumn{1}{c|}{Tasks}                         & \multicolumn{1}{c|}{Applications}                                              & \multicolumn{1}{c|}{Frame} & Representative Methods                                                                                            \\ \bottomrule

\multicolumn{1}{c|}{\multirow{5}{*}{\rotatebox{270}{~~~~~~~~~~~~~~~~~~~~Molecule Modeling}}} & \multicolumn{1}{c|}{\multirow{3}{*}{\makecell[c]{\\\\\\Molecule \\\\Conformation\\\\ Generation}}}& \multicolumn{1}{c|}{SMLD}      &    MDM~\cite{MDM}                                                                                              \\ \cline{3-4}  \multicolumn{1}{c|}{}                                    & \multicolumn{1}{c|}{}                                                             & \multicolumn{1}{c|}{DDPM}     & \begin{tabular}[c]{@{}c@{}}GeoDiff~\cite{xu2022geodiff}, \\EDMs~\cite{EDM}, \\ EEGSDE~\cite{EEGSDE}, \\DiGress~\cite{vignac2022digress}  \end{tabular}            \\ \cline{3-4} 

\multicolumn{1}{c|}{}                                    & \multicolumn{1}{c|}{}                                                  & \multicolumn{1}{c|}{SGM}      & \begin{tabular}[c]{@{}c@{}}Torsional Diffusion~\cite{jing2022torsional}, \\MOOD~\cite{MOOD}, \\GDSS~\cite{jo2022score}, \\DGSM~\cite{luo2021predicting},\\ DiffBridges~\cite{wu2022diffusionbased}\end{tabular}         \\ \cline{2-4} 
\multicolumn{1}{c|}{}                                    & \multicolumn{1}{c|}{\multirow{2}{*}{\makecell[c]{Molecular \\\\Docking}}}                & \multicolumn{1}{c|}{DDPM}     & \begin{tabular}[c]{@{}c@{}}FragDiff~\cite{FragDiff}, \\DiffLink~\cite{DiffLink}, \\ TargetDiff~\cite{Target-aware}, \\DiffBP~\cite{DiffBP}\end{tabular} \\ \cline{3-4} 
\multicolumn{1}{c|}{}                                    & \multicolumn{1}{c|}{}                                                  & \multicolumn{1}{c|}{SGM}      & DiffDock~\cite{DiffDock}                                                                                    \\ \hline 
\multicolumn{1}{c|}{\multirow{5}{*}{\rotatebox{270}{Protein Modeling}}}   & \multicolumn{1}{c|}{\multirow{3}{*}{\makecell[c]{Protein \\\\Generation}}}     & \multicolumn{1}{c|}{DDPM}     & \begin{tabular}[c]{@{}c@{}}SMCDiff~\cite{SMCDiff}, \\SiamDiff~\cite{SiamDiff}, \\ DiffFold~\cite{foldDiff},\\ ProSSDG~\cite{ProSeq},\\ DiffAntigen~\cite{luo2022antigenspecific}, \\RFdiffusion~\cite{RFDiff}\end{tabular}                       \\ \cline{3-4} 
\multicolumn{1}{c|}{}                                    & \multicolumn{1}{c|}{}                                                  & \multicolumn{1}{c|}{SGM}      & ProteinSGM~\cite{luo2022antigenspecific}                                                                                   \\ \cline{2-4}

\multicolumn{1}{c|}{}                                    & \multicolumn{1}{c|}{\multirow{2}{*}{\makecell[c]{Protein-ligand Complex  \\ Structure Prediction}}}   & \multicolumn{1}{c|}{DDPM}      & \makecell[c]{DiffEE~\cite{DiffEE}}                     \\ \cline{3-4} 
\multicolumn{1}{c|}{}                                    & \multicolumn{1}{c|}{}                                                  & \multicolumn{1}{c|}{SGM}      &     NeuralPLexer~\cite{NeuralPLexer}                                                                             \\ 
                                                    
\toprule
\end{tabular}
}
\vskip -0.15in
\end{table}

\section{Future Challenges and Opportunities}
\label{future_work}

There are increasing efforts to develop diffusion models on graphs. Next we discuss potential future research directions.     

\noindent \textbf{Discrete Nature of Graphs.} 
Most existing diffusion models for images are developed in continuous space. 
In contrast, the discrete nature of graph-structured data makes it hardly possible to directly deploy diffusion models on them. 
In this case, several works have tried to make diffusion models suitable to be used in discrete data by introducing discrete probabilistic distribution or bridging the gap between continuous and discrete spaces \cite{li2022diffusion,austin2021structured}, while there is still a lack of a universal and well-recognized method to solve this problem.

\noindent \textbf{Conditional Generation for Graph Diffusion Models}. 
Incorporating conditions into generative models is critical to guide desired generation. 
For instance, instead of generating new random samples, conditional GAN \cite{mirza2014conditional} and its variants with auxiliary knowledge have achieved remarkable success in controlling image generation. 
In graph domain, to generate molecules and proteins with specified properties, it is significant to set certain constraints on the design of graph generative models. 
Thus, introducing extra information as conditions into graph diffusion models has become an imperative research direction. 
One type of extra knowledge can be formed by a knowledge graph~\cite{chen2022knowledge}.  
Using knowledge graphs in specific fields can assist in controlling the generation process to obtain desired graphs, and enhancing the diversity of graph generation. 
In addition to knowledge graphs, other auxiliary knowledge (e.g., visual and textual) can be considered to advance the design of graph diffusion models.

\noindent \textbf{Trustworthiness for Graph Diffusion Models.}
Recent years have witnessed growing concerns about AI models' trustworthiness~\cite{liu2022trustworthy,fan2022comprehensive,fan2021jointly,fan2023adversarial,chen2023fairly}. 
As one of the most representative AI-powered applications, graph generation might cause unintentional harm to users in diverse real-world tasks, especially those in safety-critical fields such as drug discovery. 
For example, data-driven graph diffusion models are vulnerable to adversarial attacks from malicious attackers~\cite{jin2021adversarial,dai2022comprehensive};  
Due to the complexity of graph diffusion architectures, it is very challenging to understand and explain the working mechanism of graph generation~\cite{yuan2022explainability}. 
There are several crucial dimensions in achieving trustworthy graph generations, such as \emph{Safety\&Robustness, Explainability, Fairness}, and \emph{Privacy}.  
Hence, how to build trustworthy graph diffusion models has become critical in both academia and industry

\noindent \textbf{Evaluation Metrics.} 
The evaluation of graph generation remains a challenge. Most existing metrics are usually based on  graph statistics and properties (e.g., node' degree and sparsity)~\cite{o'bray2022evaluation}, which are not fully trustable.
 Meanwhile, validity and diversity for graph generation are important in different applications. 
Thus, efforts are desired to quantitatively measure the quality of generated graphs.

\noindent \textbf{Graph Diffusion Applications}. 
Most existing graph diffusion techniques are used for molecule and protein generation, while many applications on graphs are rarely explored.
\begin{list}{\labelitemi}{\leftmargin=1em}
    \setlength{\topmargin}{0pt}
    \setlength{\itemsep}{0em}
    \setlength{\parskip}{0pt}
    \setlength{\parsep}{0pt}

    \item \textbf{\textit{Recommender Systems.}} 
    The goal of recommender systems is to generate a list of  items that are likely to be clicked or purchased in the future based on users' preferences~\cite{fan2018deep,fan2021attacking,fan2022comprehensive}.  
    As users' online behaviors towards items can be naturally represented as graph-structured data,  graph learning techniques have been successfully used to capture users' preferences towards items (i.e., distributions)~\cite{wu2022graph,fan2019graph}. 
    To this end, diffusion models on graphs have the potential to model conditional distribution on items given users, so as to better generate  recommendation lists for users.

    \item \textbf{\textit{Graph Anomaly Detection.}} Anomalies are atypical data points that significantly deviate from the norm within a data distribution ~\cite{ma2021comprehensive}. In the graph domain, anomalies refer to graph objects such as nodes, edges, and sub-graphs. 
    The detection of these graph anomalies is crucial for securing against cyber attacks, detecting financial fraud, and blocking spam information. 
    Recent works have shown that diffusion models can be leveraged to purify image data for better adversarial robustness~\cite{xiao2022densepure}. 
    Thus, graph diffusion models provide great opportunities to improve graph anomaly detection, so as to enhance the graph model's robustness  against adversarial attacks.

    \item \textbf{\textit{Causal Graph Generation.}} 
    Causal Inference refers to the statistics that aim to establish the connection between cause and effect, which are usually formed by a causal-effect graph~\cite{yao2021survey}. 
    In practice, it can be difficult to analyse the relations between cause and effect because of the interference.
    For instance, instead of simply using the control {variates}, clinical trials apply causal inference to evaluate the effectiveness of the treatment. 
    In the causal discovery task, the causal-effect graph can be generated to help analyse the links between cause and effect to improve the accuracy for downstream tasks and gain explainability. 
    Therefore, graph diffusion models provide opportunities to enhance causal-effect graph generation, which can assist to reduce possible biases, build robust models, and bring new insights to explain how the model works.  
    
\end{list}

\section{Conclusion}
\label{Conclusion}

As one of the most advanced generative techniques, diffusion models have achieved great success in advancing various generative tasks, particularly in the image domain.
Similarly, many efforts have been devoted to studying graph generation based on diffusion model techniques. 
However,  it lacks a systematic overview and discussion of the state-of-the-art diffusion models on graphs.
To bridge this gap, we provided a comprehensive overview of deep diffusion models on graphs including representative models and applications. We also discussed some promising future research directions for generative diffusion models on graphs, which can bring this research field into a new frontier.

\section*{Acknowledgments}

The research described in this paper has been partly supported by NSFC (project no. 62102335),  General Research Funds from the Hong Kong Research Grants Council (Project No.: PolyU 15200021 and 15207322), internal research funds from The Hong Kong Polytechnic University (project no. P0036200 and P0042693), Collaborative Project no. P0041282, and SHTM Interdisciplinary Large Grant (project no. P0043302).
This research is also supported by the National Science Foundation (NSF) under grant numbers CNS1815636, IIS1845081, IIS1928278, IIS1955285, IIS2212032, IIS2212144, IOS2107215, and IOS2035472, the Army Research Office (ARO) under grant number W911NF-21-1-0198, the Home Depot, Cisco Systems Inc, Amazon Faculty Award, Johnson\&Johnson, and SNAP.

\balance
\bibliographystyle{ijcai19}
\bibliography{ijcai19}

\end{document}